# Active Inference for an Intelligent Agent in Autonomous Reconnaissance Missions

Johan Schubert[0000-0002-0262-9908], Farzad Kamrani[0000-0001-9448-6803], Tove Gustavi[0009-0003-1886-9265]

Swedish Defence Research Agency, SE-164 90 Stockholm, Sweden
{johan.schubert,farzad.kamrani,tove.gustavi}@foi.se

*Abstract*—We develop an active inference route-planning method for the autonomous control of intelligent agents. The aim is to reconnoiter a geographical area to maintain a common operational picture. To achieve this, we construct an evidence map that reflects our current understanding of the situation, incorporating both positive and "negative" sensor observations of possible target objects collected over time, and diffusing the evidence across the map as time progresses. The generative model of active inference uses Dempster-Shafer theory and a Gaussian sensor model, which provides input to the agent. The generative process employs a Bayesian approach to update a posterior probability distribution. We calculate the variational free energy for all positions within the area by assessing the divergence between a pignistic probability distribution of the evidence map and a posterior probability distribution of a target object based on the observations, including the level of surprise associated with receiving new observations. Using the free energy, we direct the agents' movements in a simulation by taking an incremental step toward a position that minimizes the free energy. This approach addresses the challenge of exploration and exploitation, allowing agents to balance searching extensive areas of the geographical map while tracking identified target objects.

## 1 Introduction

This paper focuses on active inference [1, 2] for the autonomous control of an intelligent agent (e.g., a reconnaissance Unmanned Aerial Vehicle (UAV)) aimed at achieving and maintaining the best possible situational awareness over time. Active inference is a methodology for autonomous decision-making. This methodology is generic and based on the concept that a system aims to minimize its surprise when receiving new information. What is unique about active inference is that the method includes two parallel approaches for an agent to seek consistency between reality and the system's description of the environment: either the system's internal representation is updated as new information is received, or actions are taken against the environment to change it so that it is consistent with its perception.

Presented at the 6th International Workshop on Active Inference, 15–17 October 2025, Montreal, Canada.

Minimizing surprise is inherently impossible. Instead, we aim for the best possible action by choosing the one that minimizes free energy. Free energy is an information theory concept that refers to the divergence between two probability distributions and the element of surprise. One distribution reflects our perception of reality, often described as a dynamic common operational picture, while the other is the probability after a current observation. When these two distributions align closely, the free energy is low. In active inference, the agent seeks to position itself so that its observations correspond with its reality model.

Section 2 formulates the problem statement. Section 3 shows related works. Section 4 introduces the fundamentals of active inference and free energy. Section 5 provides a brief overview of Dempster-Shafer theory, which updates the dynamic common operational picture. Section 6 describes the simulation environment and sensor model. In Section 7, we mathematically describe how to control an agent using active inference by minimizing free energy. Section 8 offers an overview of the implementation, and finally, Section 9 presents the conclusions drawn from this work.

## 2 Problem Statement and Persistent Surveillance

### 2.1 Problem Statement

In the field of autonomous systems, the task of continuously monitoring a specific area over an indefinite period is referred to as *persistent surveillance*. Real-world applications for persistent surveillance systems include surveillance of areas around critical infrastructure and military facilities to detect potential intruders with malicious intent. It can also be used for surveillance at popular beaches to prevent drowning incidents and for monitoring wildlife in sensitive natural areas for preservation purposes. UAVs are particularly well-suited for this type of surveillance task due to their wide-angle views, speed, and the relative absence of obstacles in their operational environments, which facilitates trajectory planning for the autonomous agents.

In this paper, we consider a scenario in which an autonomous agent is assigned the task of continuously monitoring a designated area to maintain an accurate and up-to-date dynamic common operational picture. The agent must be capable of detecting both fixed and moving targets and should ideally be able to keep track of the approximate locations of these targets, even when they are out of sensor range. Moving targets must therefore be revisited as often as necessary to prevent losing track of them. Additionally, it is requested that no part of the designated area should be left unobserved for more than a specified time.

In persistent surveillance scenarios, the trajectory planning problem for autonomous agents is usually defined by a set of specific goals and constraints, which may have different priorities. Various methods can be employed to implement the actual trajectory planning or motion control (as discussed in Section 7). This paper specifically aims to investigate the feasibility of using active inference for trajectory planning and generation. The primary research question addressed in this paper can be formulated as follows:

Can active inference be used to solve the motion control problem for a single UAV conducting multi-objective persistent surveillance over a predefined 2D area?

### 2.2 An Active Inference Approach to Persistent Surveillance

To effectively implement the active inference framework for motion control in a persistent surveillance context, several key components must be defined.

A generative model describes all possible states and their transitions to address these considerations. Information about these states is derived from observations made over time, allowing us to update our understanding of the current state. We express uncertainty in the model using Dempster-Shafer theory [3, 4]. In this model, the states are represented stochastically; each possible transition of target basic belief from one state (e.g., position) to the next is specified by transition probabilities. The states are dynamically updated at each time step using a diffusion algorithm. This model indicates the current basic belief of the existence of at least one target object within each possible state (see Section 7.1).

A generative process complements the generative model by managing new observations made with the agent's sensor. In the generative process, we use a Bayesian approach. This process updates the probabilities of all positions within the agent's sensor radius. This update calculates a new probability for each position based on the current sensor model while considering the existing probabilities (see Section 7.2).

Since we cannot directly minimize the surprise an agent experiences when receiving new observations, we focus on minimizing the free energy, which serves as an upper bound for that surprise. The free energy is defined as the sum of the divergence between two probability distributions and the level of surprise, and is our objective function to direct the agents' movements. For each location within the sensor radius, we compare the probability obtained from the generative model with the probability derived from observations, striving to minimize the divergence between the two (see Section 7.3).

At each time step, we calculate the free energy for all locations within the sensor's radius. The agent then takes a fixed-length step in the direction that minimizes free energy. This approach enables the agent to control itself autonomously, ensuring it achieves the best possible common operational picture. The agent's control is modeled by a Multi-Agent Dynamic Simulator (MADS) developed in-house.

## 3 Related Works

### 3.1 Persistent Surveillance and Motion Control

Over the years, different methods for addressing the control problem in persistent surveillance (also referred to as persistent monitoring) have been proposed. This section briefly describes some examples.

Hari et al. [5] examine a persistent monitoring mission for a single UAV, tasked with repeatedly visiting $n$ targets of equal priority. Since the targets are fixed, the problem simplifies to a classic traveling salesman problem. Brown and Anderson [6] explore a more complex maritime surveillance scenario that includes constraints on UAV dynamics and formulate a multi-objective optimization problem aimed at maximizing information gain and minimizing fuel consumption. Feasible and optimal solutions are found using a trajectory generation method combined with a particle swarm optimization algorithm. Similar to the problem discussed here (in this paper), Hübel et al. [7] introduce a dynamic "information map" subject to information decay and use it to derive a gradient-based control that drives a group of agents to continuously update their situational awareness by surveilling an area. Additionally, the authors propose a time-varying density function that can be integrated into the control algorithm to model moving points of interest. Another approach to increase the probability of autonomous agents observing moving targets during surveillance missions was proposed by Ramasamy and Ghose [8]. Assuming that the probability of observing a target is non-uniformly distributed across a monitored area, Ramasamy and Ghose assign an "importance" degree to each grid point in a discretized map where the UAV in the scenario has detected a target. The importance of a grid point depends on the number of detections made there and increases the chances of the UAV revisiting that location. Lastly, there is additional work documented in the literature that employs reinforcement learning for persistent surveillance control. Chen et al. [9] use a multi-agent reinforcement learning approach to learn policies for each agent in a team, tasked with continuously monitoring a 2D environment with stationary obstacles. To accomplish this, the problem is modeled so that a penalty is applied at every time step if a point in the environment is left unmonitored. Mishra et al. [10] present another example of a reinforcement-learning-based method for persistent surveillance.

### 3.2 Active Inference for Estimation and Control

Active inference connects perception and action through variational free energy. Theoretical links to classical estimation and control show that minimizing variational free energy yields objectives that combine information-theoretic surprise with control costs, leading to linear-quadratic-Gaussian behavior in linear-Gaussian environments [11]. In practical terms, active inference has been employed for state estimation of a quadcopter using dynamic expectation maximization (DEM). DEM is a perception scheme inspired by the brain, based on a data-driven model-learning algorithm [12]. Additionally, active inference has been used for adaptive manipulation control in the absence of detailed environment models in industrial robots. This method has proven to be scalable even when the dynamics of the environment are not explicitly modeled [13]. Active inference has also been applied to the adaptive control of robot arms using multimodal perception-action and variational autoencoder (VAE)-based state representations. This approach does not require a dynamic or kinematic model of the robot [14]. Furthermore, active inference has been used to develop a torque controller that integrates raw vision and proprioception in a streamlined design for a 7-degrees-of-freedom Franka Emika Panda robot, capable of online adaptation to changes in dynamics and human interference [15]. It has also been employed for fault-tolerant control under sensor faults, delivering unbiased state estimation and simplifying action specification [16]. These results support our use of free energy for closed-loop control with a soft sensor model.

### 3.3 Active Inference for Navigation, Exploration, and Bandit Problems

For mobile agents, active inference under hierarchical generative models enables goal-oriented navigation with topologically consistent maps and practical robot deployments [17]. Modular active inference systems support flexible, goal-driven navigation, avoiding obstacles and choosing high-confidence paths with strong zero-shot generalization to new settings [18]. Curiosity-driven learning for robotic tasks, using the free energy principle, employs Bayesian neural networks to represent epistemic uncertainty and model complex behaviors [19]. Moreover, retrospective (residual) surprise has been introduced as a computational element in active inference, serving as a lower bound on the expected free energy [20]. In decision-making, contextual multi-armed bandits (CMABs) extend the classical bandit problem by conditioning action selection on observed contextual information. Recent active inference models of CMABs [21, 22] use expected free energy to guide context-based action selection, balancing exploration and exploitation under uncertainty—an effect we replicate in our approach through minimizing a spatial free-energy field over the evidence map.

### 3.4 Multi-agent Active Inference, Organizational Adaptation, and Complex Tasks

In multi-agent settings, active inference is used to design adaptive organizations, where roles and structures change by minimizing team-level free energy [23–25]. For complex robotic tasks, active inference combines with behavior trees to enable continuous planning and robust execution with fewer nodes [26]. Meanwhile, non-modular, cognitively inspired active inference architectures are explored for robustness against unknown inputs [27]. Recent work extends active inference to autonomous driving by incorporating action-oriented priors that link perception and control, leading to more human-like, collision-avoidant behavior [28]. In socially interactive and multi-agent scenarios, active inference applies to human-robot kinesthetic interaction, where meta-priors adjust compliance and counter-forces during physical contact [29], and to empathic, socially compliant agents that adapt their movements to surrounding individuals and contextual norms [30], demonstrating the flexibility of the free-energy framework for coordination and interaction tasks. Active inference also applies to hierarchical, embodied perception-action loops, where multiple sensorimotor pathways associated with different body parts are dynamically combined, enabling the robot to reconfigure control and activate only the necessary joints for a given task [31]. Our method differs by operationalizing free energy over a spatial Dempster-Shafer evidence map and a deterministic soft sensor model, then guiding the agent to move in the direction that minimizes this per-cell objective at each step.

### 3.5 Our Contribution

Compared to surveillance controllers [5–10], we introduce a Dempster-Shafer-based evidence map with diffusion and a deterministic soft-output sensor model to generate cell-wise evidential scores. Then, we steer the platform by minimizing a grid-based free-energy objective that compares a pignistic distribution [32] to a posterior derived from the latest observation, including observation surprisal. Compared to active inference controllers and expected free energy-planning methods [13–31], our contribution is a computationally lightweight free-energy minimization over the grid map that (*i*) separates observation and state evidence, (*ii*) avoids Monte Carlo sampling by using deterministic soft observations, and (*iii*) provides an effective exploration-exploitation balance for persistent reconnaissance.

# 4 Active Inference and Free Energy

The generative model and the generative process are stochastic in nature. Let the outcome space over all possible states be $\Theta = \{T, F\}$, where $T$ denotes the presence of at least one target object, and $F$ represents the absence of a target object.

## 4.1 The Generative Model

The generative model outlines all possible states and transitions [1, 2]. In this context, the states are represented by grid cells, indicating all possible positions on a map. The agent and the target objects can transition between states, moving from any grid cell to its nearest neighboring cells.

The probability in the generative model is denoted by $q_{xy}^t(\vartheta)$, where $\vartheta$ is the state and $q_{xy}^t$ is the probability for a grid cell $c_{xy}$ at time $t$. In each grid cell, we have

$$q_{xy}^t(\vartheta = T) + q_{xy}^t(\vartheta = F) = 1. \tag{1}$$

We determine the probabilities $q_{xy}^t$ for all grid cells $c_{xy}$ in our stochastic common operational picture based on all observations gathered throughout the scenario. These observations form our understanding of the situation. More about how $q_{xy}^t$ is calculated when scouting with an agent can be found in Chapter 7.

## 4.2 The Generative Process

In the generative process, we handle new observations. At each time step, we have an incoming a priori probability $p_{xy}^t(\vartheta)$. This probability is equal to the posterior probability at the previous time step.

We set

$$p_{xy}^t(\vartheta) = p_{xy}^{t-1}(\vartheta|\varphi). \tag{2}$$

In addition, we initialize the a priori probability for time 0 according to

$$p_{xy}^0(\vartheta) = \varepsilon. \tag{3}$$

Using Bayes theorem, we can calculate the posterior probability $p_{xy}^t(\vartheta|\varphi)$ in the grid cell $c_{xy}$ for the current time step $t$ based on the a priori probability.

We have

$$p_{xy}^t(\vartheta|\varphi) = \frac{p_{xy}^t(\varphi|\vartheta)}{p_{xy}^t(\varphi)} p_{xy}^t(\vartheta), \tag{4}$$

where $p_{xy}^t(\varphi|\vartheta)$ represents the likelihood, which indicates the detection probability for an observation $\varphi = T$, as determined by the sensor model, when we have a target object $\vartheta = T$ in the grid cell $c_{xy}$ at time $t$. In this context, $p_{xy}^t(\varphi)$ represents the probability of obtaining an observation. The probability can be calculated according to

$$p_{xy}^t(\varphi) = p_{xy}^t(\varphi|\vartheta = T) \cdot p_{xy}^t(\vartheta = T) + p_{xy}^t(\varphi|\vartheta = F) \cdot p_{xy}^t(\vartheta = F), \tag{5}$$

where $p_{xy}^t(\vartheta = T)$ denotes the a priori probability discussed earlier, and

$$p_{xy}^t(\vartheta = F) = 1 - p_{xy}^t(\vartheta = T). \tag{6}$$

Additionally, $p_{xy}^t(\varphi|\vartheta = T)$ represents the likelihood, which is the detection probability of the sensor model, with

$$p_{xy}^t(\varphi|\vartheta = F) = 1 - p_{xy}^t(\varphi|\vartheta = T). \tag{7}$$

### 4.3 The Free Energy

Based on the generative model's $q_{xy}^t(\vartheta)$ and the generative process's $p_{xy}^t(\vartheta|\varphi)$, we can now calculate the free energy $F_{xy}^t$ for all grid cells $c_{xy}$ at time $t$.

We have

$$\begin{aligned} F_{xy}^t &= D_{KL}\left[q_{xy}^t(\vartheta) \parallel p_{xy}^t(\vartheta|\varphi)\right] - ln\left[p_{xy}^t(\varphi)\right] \\ &= \left[\sum_{\vartheta \in \{T,F\}} q_{xy}^t(\vartheta) \cdot ln\left(\frac{q_{xy}^t(\vartheta)}{p_{xy}^t(\vartheta|\varphi)}\right)\right] - ln\left[p_{xy}^t(\varphi)\right] \end{aligned} \tag{8}$$

here $D_{KL}$ is the Kullback-Leibler divergence [33]. The term $-ln\left[p_{xy}^t(\varphi)\right]$ represents the degree of surprise from an observation, $q_{xy}^t(\vartheta)$ is the probability according to the generative model, and $p_{xy}^t(\vartheta|\varphi)$ is the posterior probability according to the generative process, calculated at each time point based on Bayes theorem. Finally, $p_{xy}^t(\varphi)$ is the probability of an observation as derived in equation (5).

In active inference, the action that minimizes $F_{xy}^t$ is chosen. This action may involve moving to $c_{xy}$ or in its direction.

## 5 Dempster-Shafer Theory

In Dempster-Shafer theory [3, 4], belief is assigned to a proposition through a basic belief assignment. The proposition is represented by a subset $A$ of an exhaustive set of mutually exclusive possibilities, referred to as a sample space $\Theta$.

The basic belief function $m$ is defined as a function from the power set of $\Theta$ to the interval $[0, 1]$

$$m: 2^{\Theta} \rightarrow [0,1] \tag{9}$$

where $m(\emptyset) = 0$ and

$$\sum_{A \subseteq \Theta} m(A) = 1. \tag{10}$$

Here, $m(A)$ represents the basic belief assigned to $A$.

If we receive additional information about the same hypothesis from a different source, we combine the two basic belief functions to create a more comprehensive understanding. This is achieved by computing the orthogonal combination using Dempster's rule. Let $B$ be a subset of $m_1$ and $C$ be a subset of $m_2$. The combination of $m_1$ and $m_2$ results in a new basic belief function $m_1 \oplus m_2$ where

$$m_1 \oplus m_2(A) = K \cdot \sum_{B \cap C = A} m(B) \cdot m(C), \tag{11}$$

where $K$ is a normalization constant.

# 6 Simulation Environment and Sensor Model

## 6.1 Simulation Environment

The world is represented as a 2D map of a designated reconnaissance area. This map is divided into a grid of $m \times n$ square cells indexed based on a coordinate system. Using a 2D representation is a deliberate simplification justified by the fact that the scenario being examined is much larger in its $x$ and $y$ dimensions compared to any variation in its $z$ dimension. Although the simulator can perform 3D movements, this feature was disabled during the simulations conducted. The positions of both the agent and any target objects are indicated using cell indices in the format $c_{xy}$, where $x$ and $y$ are integers. Each cell is assumed to be large enough to contain the agent or a target object within the reconnaissance area.

We conduct experiments using active inference for UAV control in MATLAB. The previously mentioned 2D map is the foundation for the generative model describing the environment. The simulation environment includes both fixed and moving target objects. Two predefined movement patterns are available for the moving target objects: stationary (no movement) or movement along a straight line.

We represent uncertainty in the model using Dempster-Shafer theory [3, 4]. This uncertainty, at the map's cell level, provides two evidence-based likelihood values. The first is a basic belief, denoted as $m_{xy}^t(\vartheta = T)$, which ranges from 0 to 1. It represents the basic belief at time $t$ that the cell $c_{xy}$ contains at least one target object. The second is an estimated likelihood value, also between 0 and 1, denoted as $m_{xy}^t(\vartheta = F)$. This indicates the basic belief that, at time $t$, the cell does not contain any target objects. The formulas for diffusing basic belief to neighboring cells in the generative model are designed to be independent of the grid's topology. For example, moving from a square grid topology, used in this experiment, to a hexagonal topology only requires changing the parameter $N$, which represents the number of neighboring cells in equations (12) and (13) (see Section 7.1). This change adjusts $N$ from eight to six.

### 6.2 Sensor Model

The sensor model serves as an intelligent image sensor designed for ground scouting. Its detection radius lets it reliably identify and classify objects as targets or non-targets. Each time an object is positively classified *a*s a target, the method also provides a quantitative estimate, denoted as $m_{xy}^{t}(\vartheta = T)$, representing the basic belief that this classification is correct. If an object is identified as a target, the sensor calculates an estimate of its coordinates on a 2D map.

Since the sensor does not actively monitor or classify the *absence* of targets within the search area, it is more challenging to quantify evidence for this condition. If an area has been scanned without identified targets, it suggests no targets are present. In locations where no targets are detected, the sensor provides a default estimate, denoted as $m_{xy}^{t}(\vartheta = F)$, for the basic belief of target absence. The terrain in the area can influence this estimate; for instance, the likelihood of detecting targets is generally lower in forests compared to open fields.

We implement a simplified version of the sensor model used in the simulations to mimic its behavior as described above. The detection probability for target objects is set to 1, meaning that all target objects are detected. However, the probability that a detected target object is correctly classified is set to 0.7. Additionally, we assume the sensor does not generate false detections of target objects. Therefore, the uncertainties in the simulated sensor output arise exclusively from the classification process.

The implementation is based on the premise that a real sensor's ability to classify target objects primarily depends on the distance between the sensor and the potential target object. The sensor operates most reliably when it is focused on objects directly beneath the agent to which it is mounted. Consequently, the sensor's capability to determine whether an area is free of target objects is expected to decline as the distance from the sensor increases. In the implementation, the default values of $m_{xy}^{t}(\vartheta = T) = 0.7$ and $m_{xy}^{t}(\vartheta = F) = 0.3$ are applied. These values correspond to the maximum expected levels of $m_{xy}^{t}(\vartheta = T)$ and $m_{xy}^{t}(\vartheta = F)$ in real-world scenarios. During the simulation, these default values generate realistic outputs for $m_{xy}^{t}(\vartheta = T)$ and $m_{xy}^{t}(\vartheta = F)$ as the simulated sensor moves across the 2D map. One-dimensional Gaussian functions are used to model how these values decrease with increasing distance from the sensor.

When the sensor model is utilized in a simulation, two matrices, with basic beliefs $m_{xy}^{t}(\vartheta = T)$ and $m_{xy}^{t}(\vartheta = F)$, are generated for all grid cells $c_{xy}$ within the sensor radius using two Gaussian functions. The first Gaussian function, which produces higher values, is multiplied by a factor to set its maximum value equal to $m_{xy}^{default}(\vartheta = T) = 0.7$. This function generates $m_{xy}^{t}(\vartheta = T)$ for grid cells that contain target objects, while for these cells, $m_{xy}^{t}(\vartheta = F)$ is set to 0. The second Gaussian function is employed similarly to create $m_{xy}^{t}(\vartheta = F)$ for grid cells that do not contain target objects. This function is scaled so that its maximum value equals $m_{xy}^{default}(\vartheta = F) = 0.3$. In these grid cells, $m_{xy}^{t}(\vartheta = T)$ is set to 0. In other words, in the simulation, our sensor does not draw a stochastic categorical observation; instead, it produces a *deterministic soft output* interpreted as the expected correctness of a positive detection for each cell within the radius. We encode this soft observation as a score in [0, 1] and, for brevity, we write it using the same $m_{xy}^{t}(\bullet)$ symbols as our map-based evidence. Thus, when $m_{xy}^{t}(\vartheta = T)$ appears below in the context of the sensor, it should be read as a soft observation score produced by the sensor model, not as a posterior belief about the state.

To enhance the sensor model's realism, we introduce an assumed standard deviation, referred to as $\sigma_{position}$, which represents the error in the sensor's position estimate for identified target objects. Considering the anticipated positioning uncertainty of a sensor, the values for $m_{xy}^{t}(\vartheta = T)$ and $m_{xy}^{t}(\vartheta = F)$ in grid cells located within a distance less than the sensor's radius $r(\sigma_{position})$ are adjusted: $m_{xy}^{t}(\vartheta = T)$ is determined by the distance to the estimated target position. A Gaussian function, with an appropriate standard deviation, is applied for this calculation; $m_{xy}^{t}(\vartheta = F)$ is set equal to 0.

The Gaussian sensor model is utilized in the generative Dempster-Shafer model to quantify the observation basic belief $m_{xy}^{t}(\vartheta)$ and in the generative Bayesian process as its likelihood $p_{xy}^{t}(\varphi|\vartheta)$.

# 7 Controlling an Intelligent Agent

## 7.1 The Generative Model

This work presents the common operational picture using an evidence map, denoted as $G$. This evidence map is structured as a grid composed of multiple *cells*. Each cell corresponds to a specific segment of the geographical area being monitored, as a grid is overlaid on the map of that area. Thus, each cell represents a distinct portion of the surveillance zone.

Each cell $c_{xy}$ in the evidence map is associated with two basic beliefs: $m_{xy}^{t}(\vartheta = T)$ and $m_{xy}^{t}(\vartheta = F)$. Here, $m_{xy}^{t}(\vartheta = T)$ indicates the basic belief that the cell at time $t$ contains at least one object of interest, while $m_{xy}^{t}(\vartheta = F)$ represents the basic belief that the cell is empty. The following conditions apply to the basic beliefs: $m_{xy}^{t}(\vartheta = T), m_{xy}^{t}(\vartheta = F) \geq 0$, and $m_{xy}^{t}(\vartheta = T) + m_{xy}^{t}(\vartheta = F) \leq 1$ for all cells $c_{xy}$ at all times $t$. The basic belief for cell $c_{xy}$ at time $t$ is represented by the pair $\left(m_{xy}^{t}(\vartheta = T), m_{xy}^{t}(\vartheta = F)\right)$, with uncertainty defined as $1 - m_{xy}^{t}(\vartheta = T) - m_{xy}^{t}(\vartheta = F)$.

In Fig. 1, the green area represents $m_{xy}^{t}(\vartheta = T)$, the red area represents $m_{xy}^{t}(\vartheta = F)$, and the white area indicates the uncertainty.

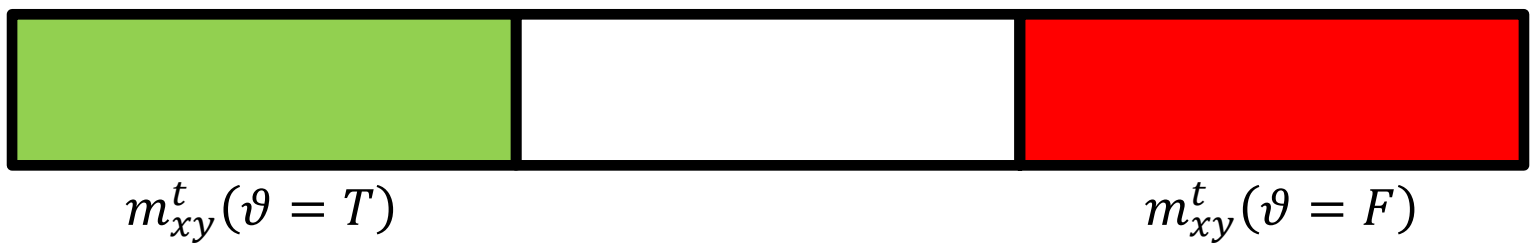

**Fig 1.** Illustration of evidence and uncertainty in cell $c_{xy}$. Green indicates $m_{xy}^t(\vartheta = T)$, red indicates $m_{xy}^t(\vartheta = F)$, and white shows residual uncertainty $1 - m_{xy}^t(\vartheta = T) - m_{xy}^t(\vartheta = F)$; this per-cell Dempster–Shafer pair forms the basis of the evidence map $G$.

The uncertainty $1 - m_{xy}^t(\vartheta = T) - m_{xy}^t(\vartheta = F)$ can be considered a genuine uncertainty in cell $c_{xy}$ that the agent should strive to minimize across all cells $c_{xy}$ and at all times to maintain a current common operational picture.

Instead of assigning a single piece of evidence for a positive detection to one specific set of grid cells within the assumed detection radius of the sensors, we opt to assign evidence to each grid cell individually within that radius. This approach offers a more detailed representation, as each grid cell $c_{xy}$ receives its value, while significantly reducing the computational complexity involved in updating these values with new sensor measurements.

Our experiment utilizes a Gaussian distribution with a standard deviation equal to the sensor radius, ensuring that the center cell is assigned the highest value. This method results in a series of distinct basic belief distributions – one for each grid cell. The outcome provides a satisfactory resolution and a relevant spread in the evidence-based common operational picture. The diffusion model outlined in equations (12) and (13) achieves a dynamic evidence-based common operational picture.

Given that the observed environment is assumed to be dynamic, the value of older information diminishes over time. In the evidence map, this decreasing certainty about the locations of already detected objects is represented by the spread of basic belief to adjacent cells.

Regardless of whether the agent observed the state of cell $c_{xy}$ between times $t$ and $t + 1$, the basic beliefs $m_{xy}^t(\vartheta = T)$ and $m_{xy}^t(\vartheta = F)$ in the cell are updated according to equations (12) and (13). This is separate from fusion with new observations, which is managed in equations (14) and (15). The update for the basic belief is given by:

$$m_{xy}^{t+1}(\vartheta = T) = 1 - \left(1 - \frac{m_{xy}^t(\vartheta = T)}{N + 1}\right) \cdot \prod_{(ij) \in \{neighbors\ of\ c_{xy}\}} \left(1 - \frac{m_{ij}^t(\vartheta = T)}{N + 1}\right) \tag{12}$$

where $N$ is the number of neighbors of the cell $c_{xy}$. This means that the basic belief $m_{xy}^t(\vartheta = T)$ of cell $c_{xy}$ is shared among its neighbors and itself (i.e., among nine grid cells in the case of a square grid) and then combined using Dempster's rule. This causes a diffusion of basic belief $m_{ij}^t(\vartheta = T)$ over an increasingly larger area over time (in a manner similar to an explosion).

Suppose $m_{xy}^{t}(\vartheta = T) > 0$ in cell $c_{xy}$, some of the basic belief $m_{xy}^{t}(\vartheta = T)$ will be propagated to its $N$ neighbors. Conversely, if $m_{ij}^{t}(\vartheta = T) > 0$ in adjacent cells, some basic belief is propagated into cell $c_{xy}$.

Furthermore, we have

$$m_{xy}^{t+1}(\vartheta = F) = m_{xy}^{t}{}^{\left(\frac{1}{N+1}\right)}(\vartheta = F) \cdot \prod_{(ij) \in \{neighbors\ of\ c_{xy}\}} m_{ij}^{t}{}^{\left(\frac{1}{N+1}\right)}(\vartheta = F) \quad (13)$$

where $m_{xy}^{t+1}(\vartheta = F)$ indicates that cell $c_{xy}$ is empty at time $t + 1$. This suggests that if cell $c_{xy}$ will be empty at the next time step (i.e., $m_{xy}^{t+1}(\vartheta = F)$), then both its neighboring cells and the cell itself must currently be empty, meaning $m_{xy}^{t}(\vartheta = F)$. Equation (13) contracts basic belief $m_{xy}^{t}(\vartheta = F)$ inward (in a manner similar to an implosion).

A sensor on an agent gathers information about one or more cells in the evidence map. This gathering is influenced by the sensor's detection radius and the agent's flight altitude. As the flight altitude rises, the likelihood of detecting interesting objects on the ground decreases. In our simulations, we keep a constant flight altitude since we only model the problem in 2D.

A positive observation occurs when the sensor detects an object of interest in a specific cell $c_{xy}$ at time $t$, represented as $\left(m_{xy}^{t+1}(\varphi = T), 0\right)$, where $m_{xy}^{t+1}(\varphi = T)$ indicates the basic belief that the positive detection is correct. Conversely, a "negative" observation, where the sensor does not detect any object of interest in cell $c_{xy}$ at time $t$, is represented as $\left(0, m_{xy}^{t+1}(\varphi = F)\right)$. Here, $m_{xy}^{t+1}(\varphi = F)$ represents the basic belief confirming that the negative result is accurate.

Both $m_{xy}^{t+1}(\varphi = T)$ and $m_{xy}^{t+1}(\varphi = F)$ fall within the range of $[0,1]$. Sensor measurements continuously update the evidence map according to equations (14) and (15) below.

If we obtain a positive observation $\left(m_{xy}^{t+1}(\varphi = T), 0\right)$, it is combined with the existing basic belief pair $m_{xy}^{t+1}(\vartheta = T)$ and $m_{xy}^{t+1}(\vartheta = F)$, which are derived using equations (13) and (14) with Dempster's rule [3].

We have

$$m_{xy}^{t+1*}(\vartheta = T) = \\ = \frac{m_{xy}^{t+1}(\vartheta = T) + m_{xy}^{t+1}(\varphi = T) \cdot \left(1 - m_{xy}^{t+1}(\vartheta = T) - m_{xy}^{t+1}(\vartheta = F)\right)}{1 - m_{xy}^{t+1}(\varphi = T) \cdot m_{xy}^{t+1}(\vartheta = F)} \quad (14)$$

and

$$m_{xy}^{t+1*}(\vartheta = F) = \frac{\left(1 - m_{xy}^{t+1}(\varphi = T)\right) \cdot m_{xy}^{t+1}(\vartheta = F)}{1 - m_{xy}^{t+1}(\varphi = T) \cdot m_{xy}^{t+1}(\vartheta = F)}. \quad (15)$$

It is assumed that equations (14) and (15) are always applied after equations (12) and (13), and at the same time, $t$. The basic beliefs $m_{xy}^{t+1}(\vartheta = T)$ and $m_{xy}^{t+1}(\vartheta = F)$ on the right-hand side represent the results from the diffusion update at time step $t + 1$. In contrast, the values $m_{xy}^{t+1*}(\vartheta = T)$ and $m_{xy}^{t+1*}(\vartheta = F)$ on the left-hand side, indicated with an asterisk (*), are the combined results from the two updates at the same time step $t + 1$.

If a new "negative" observation is received in grid cell $c_{xy}$, then $\left(0, m_{xy}^{t+1}(\varphi = F)\right)$ is combined with $m_{xy}^{t+1}(\vartheta = T)$ and $m_{xy}^{t+1}(\vartheta = F)$ using Dempster's rule.

We have

$$m_{xy}^{t+1*}(\vartheta = T) = \frac{m_{xy}^{t+1}(\vartheta = T) \cdot \left(1 - m_{xy}^{t+1}(\varphi = F)\right)}{1 - m_{xy}^{t+1}(\vartheta = T) \cdot m_{xy}^{t+1}(\varphi = F)} \tag{16}$$

and

$$m_{xy}^{t+1*}(\vartheta = F) =$$

$$= \frac{m_{xy}^{t+1}(\vartheta = F) + m_{xy}^{t+1}(\varphi = F) \cdot \left(1 - m_{xy}^{t+1}(\vartheta = T) - m_{xy}^{t+1}(\vartheta = F)\right)}{1 - m_{xy}^{t+1}(\vartheta = T) \cdot m_{xy}^{t+1}(\varphi = F)}. \tag{17}$$

The result of the distribution $m_{xy}^{t+1*}(\vartheta = T)$ based on equations (14) and (16) is shown in Fig. 2. Here, the result in the figure shows the basic belief from positive and "negative" observations.

### 7.2 The Generative Process

In the generative process, we employ a Bayesian approach. When we receive a current observation, our goal is to calculate the updated posterior distribution $p_{xy}^{t+1}(\vartheta|\varphi)$. We must determine this probability distribution across all grid cells, $c_{xy}$, to achieve this. This requires a sensor model in the form of a likelihood distribution $p_{xy}^{t+1}(\varphi|\vartheta)$ and an observation model $p_{xy}^{t+1}(\varphi)$, as well as an a priori distribution $p_{xy}^{t+1}(\vartheta)$.

We begin with the a priori distribution. It is important to note that the a priori distribution at each time step equals the posterior distribution from the previous time point.

We have

$$p_{xy}^{t+1}(\vartheta) = p_{xy}^{t}(\vartheta|\varphi). \tag{18}$$

Initially, we have set

$$p_{xy}^{0}(\vartheta) = \varepsilon = \frac{\#\ assumed\ targets\ in\ the\ current\ area}{\#\ grid\ cells\ in\ the\ current\ area}. \tag{19}$$

Using a likelihood distribution and our sensor model, we assign probabilities to each grid cell $c_{xy}$ within the sensor's radius based on two Gaussian distributions.

To calculate the probability of an observation, denoted as $p_{xy}^{t}(\varphi)$, we can express it as follows:

$$p_{xy}^{t+1}(\varphi) = p_{xy}^{t+1}(\varphi|\vartheta = T) \cdot p_{xy}^{t+1}(\vartheta = T) + p_{xy}^{t+1}(\varphi|\vartheta = F) \cdot p_{xy}^{t+1}(\vartheta = F), \tag{20}$$

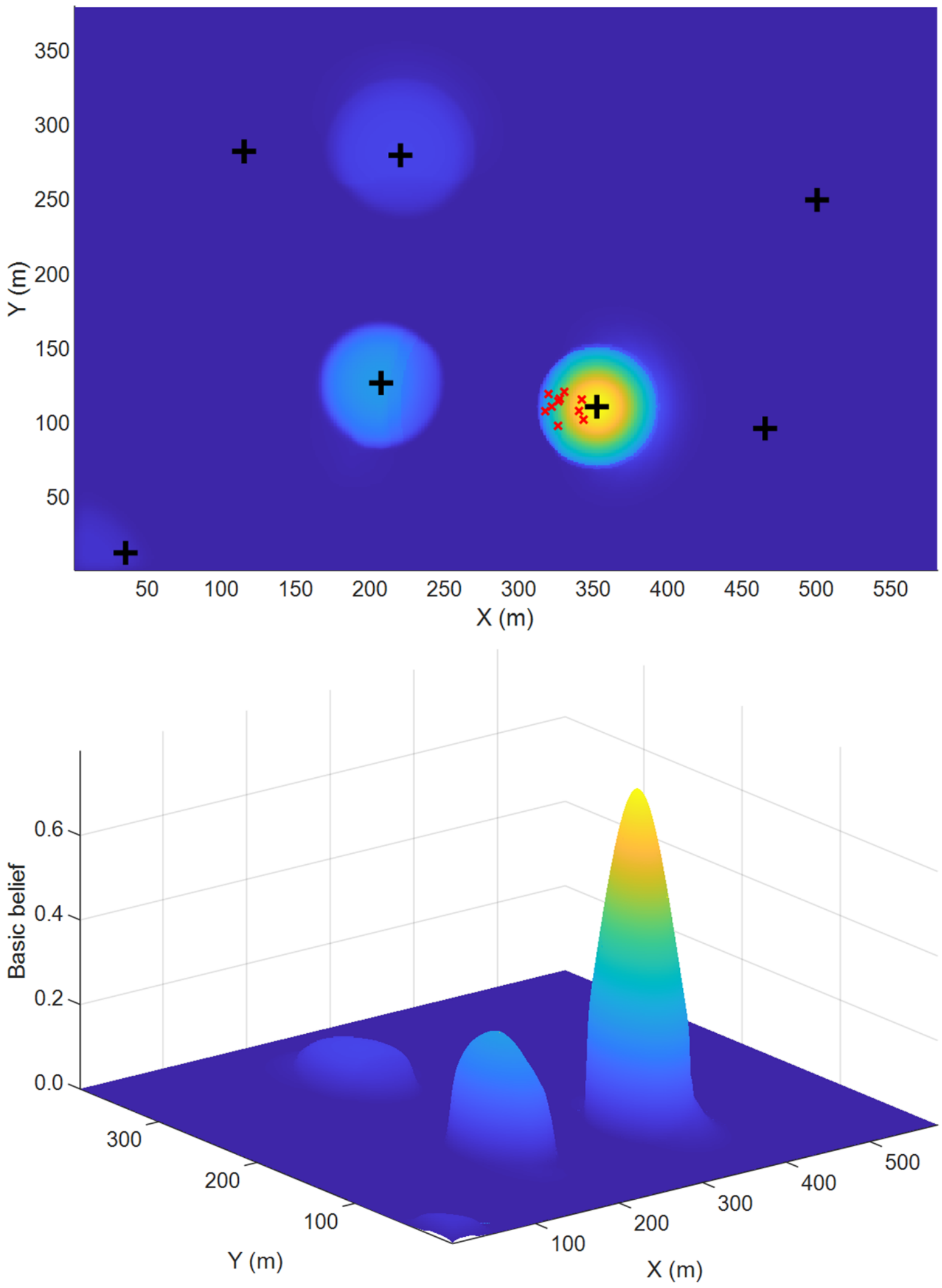

**Fig. 2.** Fused results, $m_{xy}^{t+1*}(\vartheta = T)$ derived from the generative model in a simulation scenario at 50 seconds (2D on top and 3D below). Values reflect diffusion, equations (12)–(13), followed by Dempster–Shafer fusion, equations (14)–(17); higher values indicate a stronger basic belief that a cell contains ≥ 1 target (axes in meters; black "+" marks show target locations).

where $p_{xy}^{t+1}(\vartheta = T)$ is the a priori probability derived earlier in (18) and

$$p_{xy}^{t+1}(\vartheta = F) = 1 - p_{xy}^{t+1}(\vartheta = T). \qquad (21)$$

Additionally, the likelihood distribution $p_{xy}^{t}(\varphi|\vartheta = T)$ is calculated using a Gaussian distribution, similarly to the method described in Section 6.2, starting from $m_{xy}^{default}(\vartheta = T)$.

Since the sensor returns a deterministic score, we define the likelihood directly from it. Hence, the identification in (22) is just shorthand rather than an inference about the state,

$$p_{xy}^{t+1}(\varphi|\vartheta = T) \triangleq m_{xy}^{t+1}(\vartheta = T),$$
$$p_{xy}^{t+1}(\varphi|\vartheta = F) = 1 - p_{xy}^{t+1}(\varphi|\vartheta = T). \quad (22)$$

Using the three probability distributions (a priori $p_{xy}^{t+1}(\vartheta)$, likelihood $p_{xy}^{t+1}(\varphi|\vartheta)$, and observation $p_{xy}^{t+1}(\varphi)$), we can calculate the desired posterior distribution using Bayes theorem.

We have

$$p_{xy}^{t+1}(\vartheta|\varphi) = \frac{p_{xy}^{t+1}(\varphi|\vartheta)}{p_{xy}^{t+1}(\varphi)} p_{xy}^{t+1}(\vartheta). \quad (23)$$

Fig. 3 displays the posterior distribution concurrently with Fig. 2.

### 7.3 The Free Energy

The basic belief distribution $m_{xy}^{t+1*}(\vartheta = T)$ represents the distribution from the generative model that we aim to compare with the posterior distribution derived from the current observation $p_{xy}^{t+1}(\vartheta|\varphi)$. Notably, $m_{xy}^{t+1*}(\vartheta = T)$ encompasses values over all subsets of the sample space $\Theta$, including $\Theta$ itself (where $|2^{\Theta}| = 3$), while $p_{xy}^{t}(\vartheta|\varphi)$ is limited to values associated with the elements of $\Theta$ (where $|\Theta| = 2$). Therefore, we must convert the basic belief distribution into a probability distribution to facilitate a comparison between these two distributions. This conversion is achieved through a pignistic transformation [32], defined as follows:

$$BetP(\omega) = \sum_{\omega \in X} \frac{m(X)}{|X|} \quad (24)$$

where $\omega \in \Theta$ and $X \subseteq \Theta$.

We have

$$p_{xy}^{t+1*}(\vartheta = T) = \frac{1}{2}\left[1 + m_{xy}^{t+1*}(\vartheta = T) - m_{xy}^{t+1*}(\vartheta = F)\right] \quad (25)$$

and

$$p_{xy}^{t+1*}(\vartheta = F) = \frac{1}{2}\left[1 + m_{xy}^{t+1*}(\vartheta = F) - m_{xy}^{t+1*}(\vartheta = T)\right], \quad (26)$$

where $p_{xy}^{t+1*}(\vartheta)$ represents a pignistic probability derived from the pignistic transformation of the basic belief distribution $m_{xy}^{t+1*}(\vartheta)$.

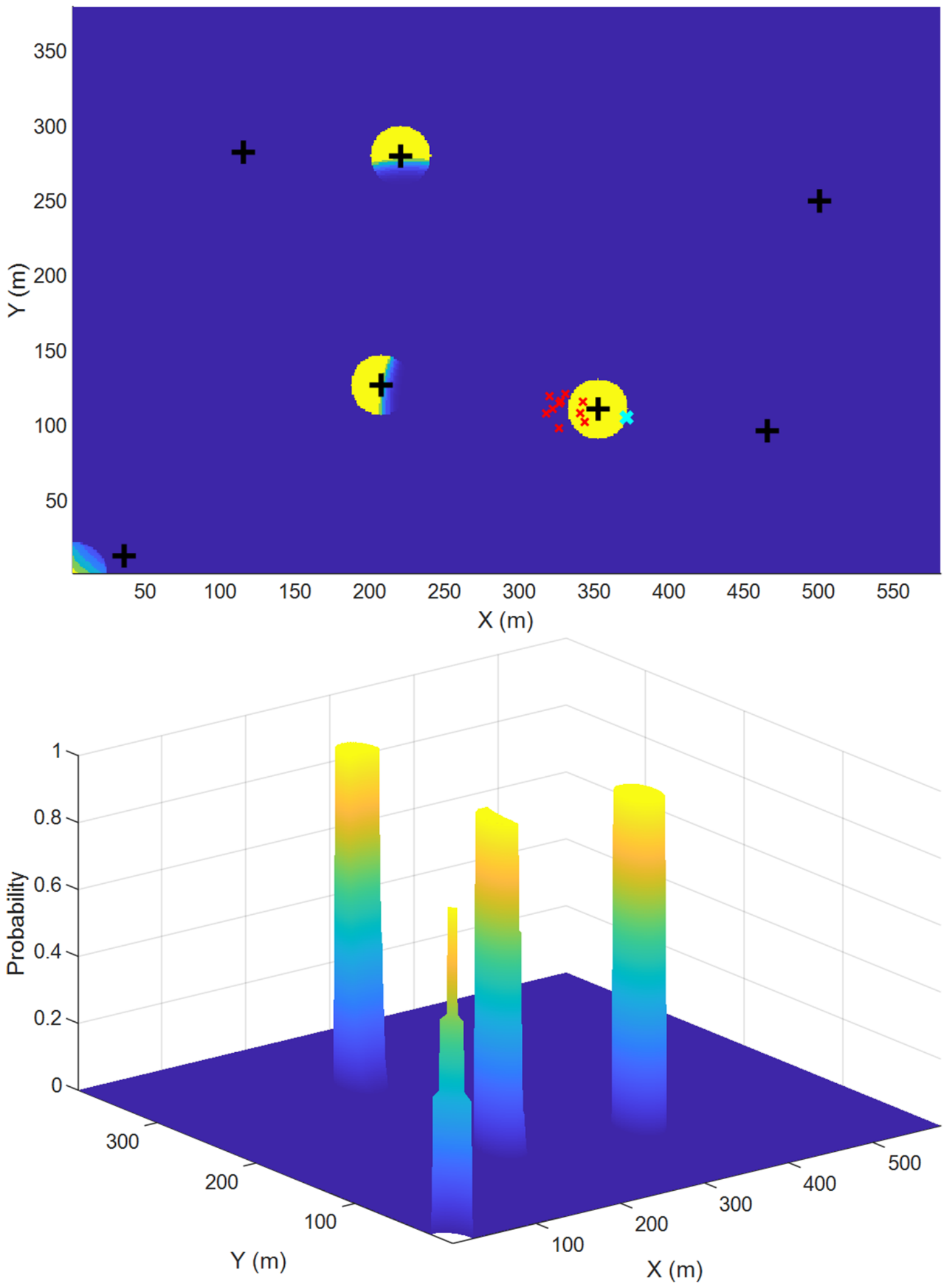

**Fig. 3.** The posterior distribution $p_{xy}^{t+1}(\vartheta|\varphi)$, calculated after the most recent observation, using Bayes rule, equation (23), with the prior, equations (18)–(21), and Gaussian sensor likelihood, equation (22), at the same timestep as Fig. 2, for comparison with the fused belief shown there.

We can now calculate the free energy of each grid cell $c_{xy}$ within the sensor radius by measuring the divergence between $p_{xy}^{t+1*}(\vartheta)$ and $p_{xy}^{t+1}(\vartheta|\varphi)$, along with the degree of surprise represented by $p_{xy}^{t+1}(\varphi)$.

Fig. 4 illustrates the divergence between the pignistic probability distribution $p_{xy}^{t+1*}(\vartheta)$ from the generative model and the posterior distribution $p_{xy}^{t+1}(\vartheta|\varphi)$ from the generative process.

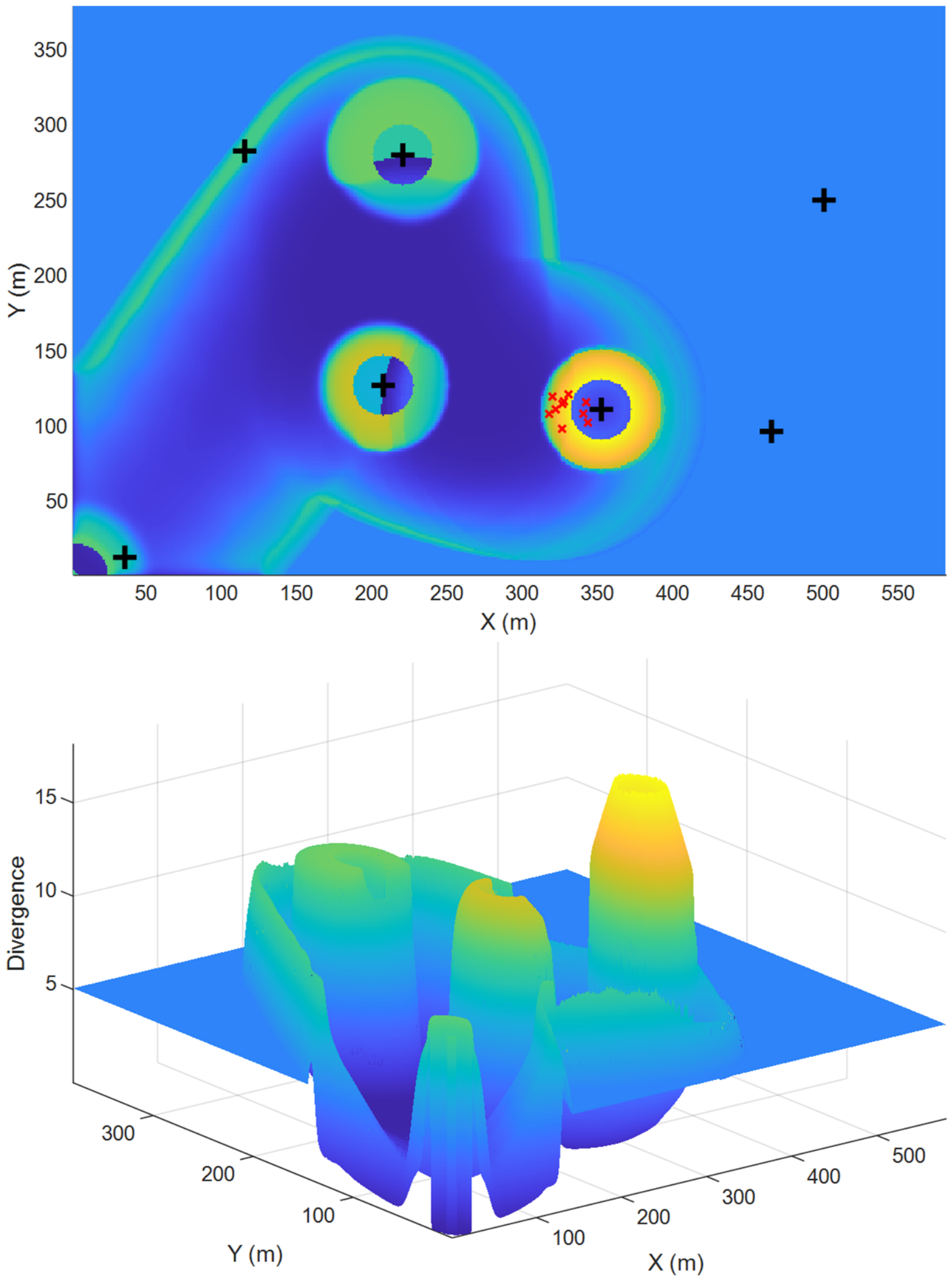

**Fig. 4.** The divergence between the model and the actual observation. Cell-wise $D_{KL}\left[p_{xy}^{t+1*}(\vartheta) \parallel p_{xy}^{t+1}(\vartheta|\varphi)\right]$, which is the first term of equation (27) within the sensor footprint ($r$ = 100 meters), illustrating where the model's belief and the posterior disagree.

Fig. 5 displays the degree of surprise, indicated as $-ln\left[p_{xy}^{t+1}(\varphi)\right]$.

We have

$$F_{xy} = D_{KL}\left[p_{xy}^{t+1*}(\vartheta) \parallel p_{xy}^{t+1}(\vartheta|\varphi)\right] - ln\left[p_{xy}^{t+1}(\varphi)\right]$$

$$= \left[\sum_{\vartheta\in\{T,F\}} p_{xy}^{t+1*}(\vartheta) \cdot ln\left(\frac{p_{xy}^{t+1*}(\vartheta)}{p_{xy}^{t+1}(\vartheta|\varphi)}\right)\right] - ln\left[p_{xy}^{t+1}(\varphi)\right], \tag{27}$$

where $D_{KL}$ is the Kullback-Leibler divergence [33].

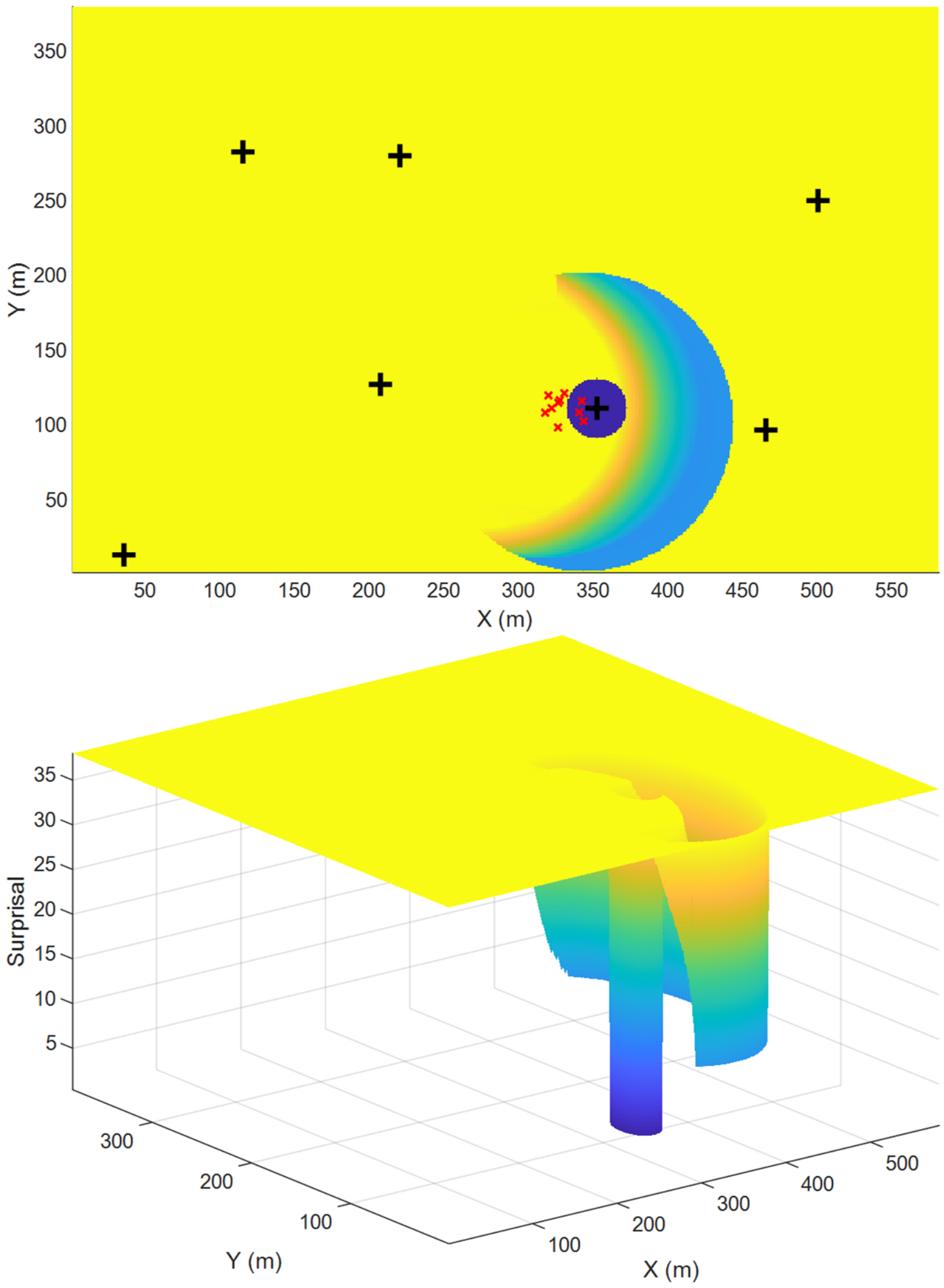

**Fig. 5.** The level of surprise. Surprisal $-ln\left[p_{xy}^{t+1}(\varphi)\right]$, which represents the second term of equation (27), is shown inside the sensor footprint. Outside the footprint, $p(\varphi) = 0$, so surprisal diverges and is omitted.

The free energy $F_{xy}$ is calculated for all grid cells within the sensor radius and minimized across all grid cells. Finally, the variational free energy is presented in Fig. 6.

The agent's position is updated with steps smaller than the sensor radius. Thus, using active inference, we gradually move the agent's position toward the grid cell, which minimizes $F_{xy}$.

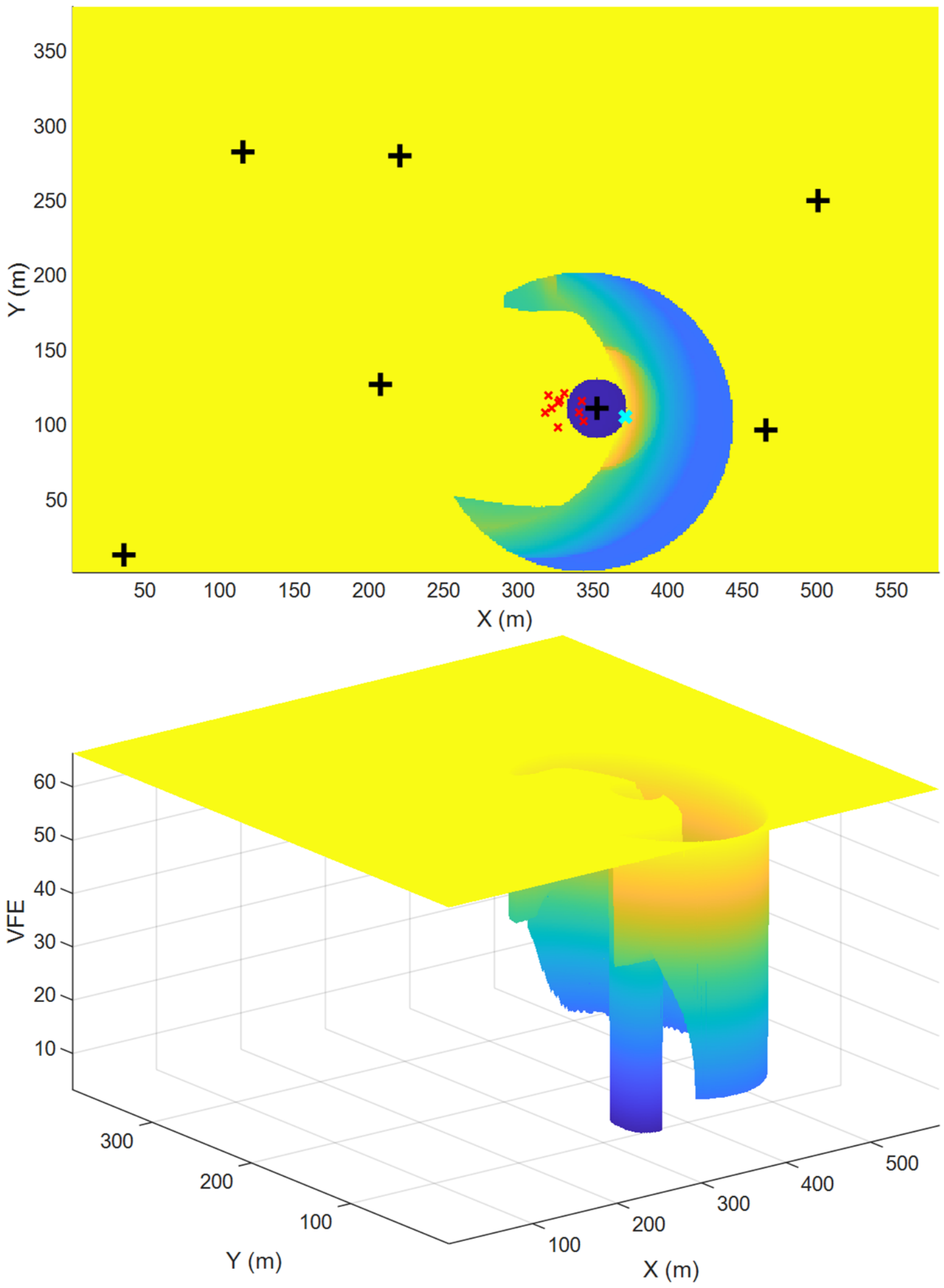

**Fig. 6.** Free energy, denoted as $F_{xy}$, for all grid cells $c_{xy}$. $F_{xy}$ is calculated using equation (27); the green "×" marks argmin $F$, which the agent uses to set the next waypoint.

## 8 Implementation and Results for an Intelligent Agent

Implementing agent control using the free energy principle and active inference involves calculations, as outlined in Section 7. However, several essential implementation details and design choices must be considered.

The generative model and process are evaluated at each simulation time step. Importantly, the free energy is calculated only for positions within the sensor radius; for this simulation, we use a sensor radius of 100 meters. This limitation is not merely a design choice but a fundamental aspect of the model. According to equation (27), free energy is defined as the sum of the Kullback-Leibler divergence between two probabilities, $D_{KL}\left[p_{xy}^{t+1*}(\vartheta = T) \parallel p_{xy}^{t+1}(\vartheta|\varphi)\right]$, and the surprise, which is calculated as $-ln\left[p_{xy}^{t+1}(\varphi)\right]$. Here, $p_{xy}^{t+1}(\varphi)$ represents the probability of the observation. Outside the sensor radius, $p_{xy}^{t+1}(\varphi)$ equals zero, causing $-ln\left[p_{xy}^{t+1}(\varphi)\right]$ to approach infinity. This observation highlights the notion that surprise becomes infinite when faced with impossibilities.

The computational complexity per step of our planner is

$$O\left(N_{agents}^2 + N_{poi}A_g + A_g + s_r^2\right), \tag{28}$$

where $N_{agents}$ is the number of agents in the swarm, $A_g$ is the number of grid cells, $N_{poi}$ is the number of points of interest (POI), and $s_r$ is the sensing radius expressed in grid cells. In practice, for a $580 \times 380$ grid with ten agents, six POI, and a sensing radius of $100$ cells (i.e., $100$ meters), each simulation step took on average $0.12$ seconds, which amounts to 36 seconds for a 300-step rollout (one minute wall-clock time at five simulation ticks per second, i.e., faster than real time) on a standard laptop (MATLAB; Intel i7; 32 GB RAM).

Another design choice focuses on how to select the new waypoint. Once the position with the lowest free energy has been identified, we calculate a vector from the agent's current position to the minimum free energy position. The agent then takes an incremental step toward that position. The next waypoint is placed at a fixed distance along this vector, which in our simulation is set to half the sensor radius (i.e., a step length of $50$ meters). This approach ensures a consistent distance between successive waypoints.

This section provides a qualitative analysis of the method's behavior based on simulation observations. The findings indicate that the technique demonstrates favorable behavior. After navigating the pre-programmed waypoints (the first three targets), the agent continues to search the area in a balanced manner that integrates *exploration* and *exploitation*.

An agent's trajectory (shown as a red line) over a 500-second scenario (Fig. 7) illustrates a dynamic balance between tracking various target goals (indicated by black "**+**" symbols) and exploring new areas. Black trajectories represent three moving targets, while four targets are stationary. The green "**x**" indicates the minimum free energy at the current time.

A qualitative analysis of the images using a stochastic model, along with probabilistic observations and their divergences, provides insights into the level of surprise experienced by the agent. This understanding and the concept of free energy help clarify the reasons and timing behind the agent's decision to scout or pursue a target. When examining the sequence of images throughout a scenario, a behavior that is intuitively and analytically coherent emerges.

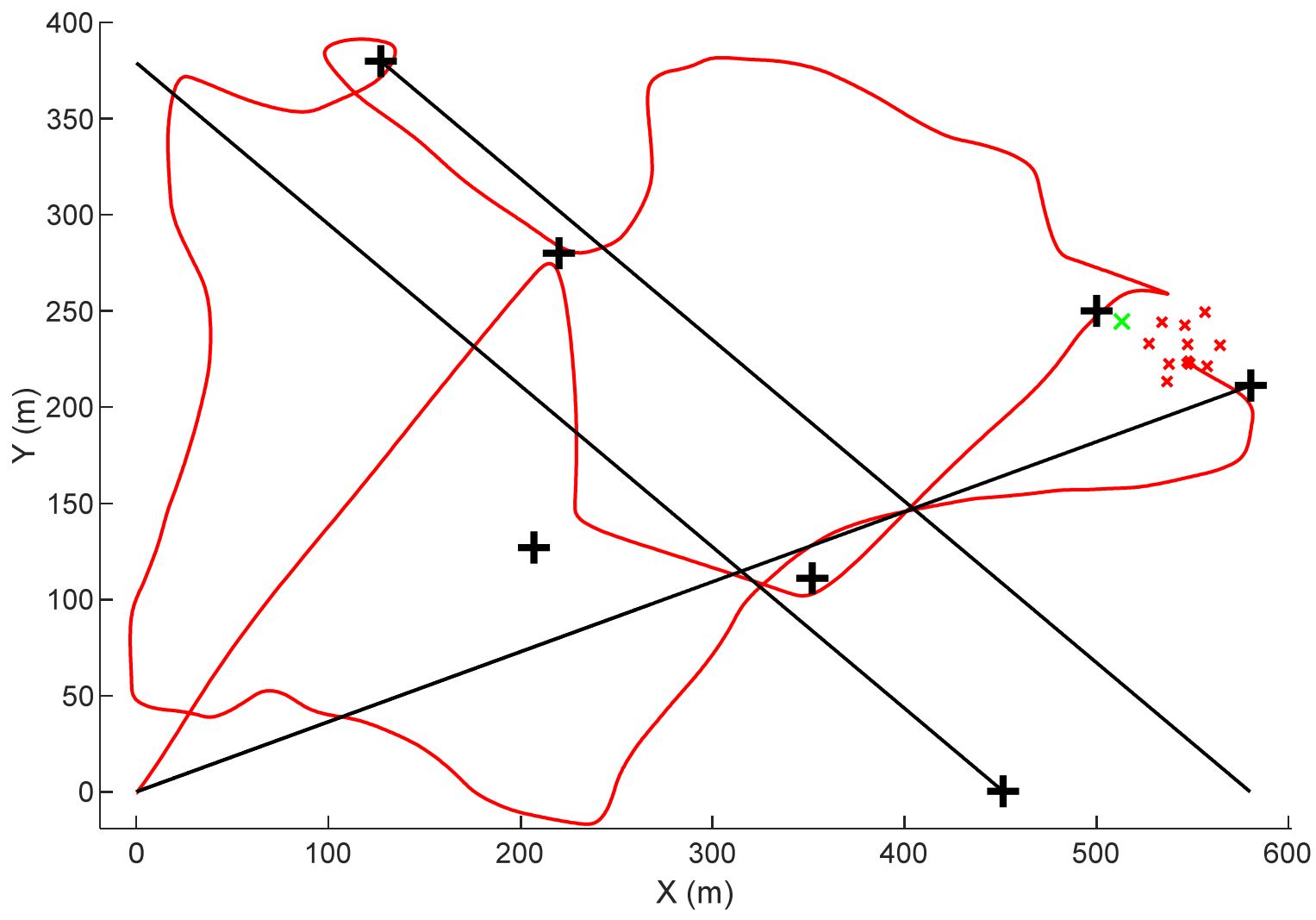

**Fig. 7.** The agent's path entails both reconnaissance and target tracking. The trajectory, waypoints, and position of minimum free energy are illustrated. The agents, represented by red crosses, select a new waypoint along a vector that points towards the minimum free energy point, indicated by the green cross. The black crosses depict the exact locations of the targets from a global perspective; these locations remain unknown to the agents except through sensor observations. The black "**+**" at the end of black trajectories indicates the final position of a target as it exits the map and the simulation.

## 9 Conclusions

A simulation of an intelligent agent demonstrates its ability to autonomously decide when to engage in scouting activities and when to track specific targets. This decision-making process entirely depends on the data acquired through the agent's sensors. As the agent operates, it navigates from its current location towards the designated grid cell that minimizes free energy. Free energy quantifies the degree of alignment between the agent's perception of reality and its current observations. The agent's minimizing free energy enhances its comprehension and adaptability to the external environment. Additionally, the planner operates efficiently: in our prototype, each step takes approximately 0.12 seconds on average, resulting in 36 seconds for a 300-step rollout using standard hardware, which demonstrates its potential for near real-time operation. A qualitative analysis of the agent's behavior during the simulation reveals promising results. The agent's ability to balance *exploration* and *exploitation* indicates significant operational autonomy and efficiency. A quantitative analysis of the active inference approach to the persistent monitoring problem has not yet been performed. To achieve this, some mission goals need to be formulated, and appropriate metrics for measuring goal fulfillment must be defined. Examples of performance metrics for a combined tracking and area coverage scenario, similar to the scenario discussed in this paper, can be found, for instance, in [8]. The performance of the implemented active inference-based control should be compared to that of a simple baseline control or, ideally, with several other control algorithms from the literature, and for some different scenarios.

**Disclosure of Interests.** The authors have no competing interests to declare relevant to this article's content.

## References

1. Parr, T., Pezzulo, G., Friston, K. J.: Active inference: The free energy principle in mind, brain, and behavior. The MIT Press, Cambridge, MA (2022). doi:10.7551/mitpress/12441.001.0001
2. Buckley, C. L., Kim, C. S., McGregor, S., Seth, A. K.: The free energy principle for action and perception: A mathematical review. Journal of Mathematical Psychology **81**, 55–79 (2017). doi:10.1016/j.jmp.2017.09.004
3. Dempster, A. P.: A generalization of Bayesian inference. Journal of the Royal Statistical Society. Series B **30**(2), 205–247 (1968). doi:10.1111/j.2517-6161.1968.tb00722.x
4. Shafer, G.: A mathematical theory of evidence. Princeton University Press, Princeton, NJ (1976)
5. Hari, S. K. K., Rathinam, S., Darbha, S., Kalyanam, K., Manyam, S. G., Casbeer, D.: Optimal UAV route planning for persistent monitoring missions. IEEE Transactions on Robotics **37**(2), 550–566 (2020). doi:10.1109/TRO.2020.3032171
6. Brown, A., Anderson, D.: Trajectory optimization for high-altitude long-endurance UAV maritime radar surveillance. IEEE Transactions on Aerospace and Electronic Systems **56**(3), 2406–2421 (2020). doi:10.1109/TAES.2019.2949384
7. Hübel, N., Hirche, S., Gusrialdi, A., Hatanaka, T., Fujita, M., Sawodny, O.: Coverage control with information decay in dynamic environments. IFAC Proceedings **41**(2), 4180–4185 (2008). doi:10.3182/20080706-5-KR-1001.00703
8. Ramasamy, M., Ghose, D.: Learning-based preferential surveillance algorithm for persistent surveillance by unmanned aerial vehicles. In: Proceedings of the 2016 International Conference on Unmanned Aircraft Systems, pp. 1032–1040. IEEE, Piscataway, NJ (2016). doi:10.1109/ICUAS.2016.7502678
9. Chen, J., Baskaran, A., Zhang, Z., Tokekar, P.: Multi-agent reinforcement learning for visibility-based persistent monitoring. In: Proceedings of the 2021 IEEE/RSJ International Conference on Intelligent Robots and Systems, pp. 2563–2570. IEEE, Piscataway, NJ (2021). doi:10.1109/IROS51168.2021.9635898
10. Mishra, M., Poddar, P., Agrawal, R., Chen, J., Tokekar, P., Sujit, P. B.: Multi-agent deep reinforcement learning for persistent monitoring with sensing, communication, and localization constraints. IEEE Transactions on Automation Science and Engineering **22**, 2831–2843 (2025). doi:10.1109/TASE.2024.3385412
11. van de Laar, T., Özçelikkale, A., Wymeersch, H.: Application of the free energy principle to estimation and control. IEEE Transactions on Signal Processing **69**, 4234–4244, 2021. doi:10.1109/TSP.2021.3095711
12. Bos, F., Meera, A. A., Benders, D., Wisse, M.: Free energy principle for state and input estimation of a quadcopter flying in wind. In: Proceedings of the 2022 International Conference on Robotics and Automation, pp. 5389–5395. IEEE, Piscataway, NJ (2022). doi:10.1109/ICRA46639.2022.9812415

13. Pezzato, C., Ferrari, R., Hernández Corbato, C.: A novel adaptive controller for robot manipulators based on active inference. IEEE Robotics and Automation Letters **5**(2), 2973–2980 (2020). doi:10.1109/LRA.2020.2974451
14. Meo, C., Lanillos, P.: Multimodal VAE active inference controller. In: Proceedings of the 2021 IEEE/RSJ International Conference on Intelligent Robots and Systems, pp. 2693–2699. IEEE, Piscataway, NJ (2021). doi:10.1109/IROS51168.2021.9636394
15. Meo, C., Franzese, G., Pezzato, C., Spahn, M., Lanillos, P.: Adaptation through prediction: multisensory active inference torque control. IEEE Transactions on Cognitive and Developmental Systems **15**(1), 32–41 (2023). doi:10.1109/TCDS.2022.3156664
16. Baioumy, M., Pezzato, C., Ferrari, R., Corbato, C. H., Hawes, N.: Fault-tolerant control of robot manipulators with sensory faults using unbiased active inference. In: Proceedings of the 2021 European Control Conference, pp. 1119–1125. IEEE, Piscataway, NJ (2021). doi:10.23919/ECC54610.2021.9654913
17. Çatal, O., Verbelen, T., Van de Maele, T., Dhoedt, B., Safron, A.: Robot navigation as hierarchical active inference. Neural Networks **142**, 192–204 (2021). doi:10.1016/j.neunet.2021.05.010
18. Scholz, F., Gumbsch, C., Otte, S., Butz, M. V.: Inference of affordances and active motor control in simulated agents. Frontiers in Neurorobotics 16, 881673 (2022). doi:10.3389/fnbot.2022.881673
19. Kawahara, D., Ozeki, S., Mizuuchi, I.: A curiosity algorithm for robots based on the free energy principle. In: Proceedings of the 2022 IEEE/SICE International Symposium on System Integration, pp. 53–59. IEEE, Piscataway, NJ (2022). doi:10.1109/SII52469.2022.9708819
20. Katahira, K., Kunisato, Y., Okimura, T., Yamashita, Y.: Retrospective surprise: A computational component for active inference. Journal of Mathematical Psychology **96**, 102347 (2020). doi:10.1016/j.jmp.2020.102347
21. Wakayama, S., Ahmed, N.: Active Inference for autonomous decision-making with contextual multi-armed bandits. In: Proceedings of the 2023 IEEE International Conference on Robotics and Automation, pp. 7916–7922 (2023). doi: 10.1109/ICRA48891.2023.10160593
22. Wakayama, S., Candela, A., Hayne, P., Ahmed, N.: Active inference for bandit-based autonomous robotic exploration with dynamic preferences. IEEE Transactions on Robotics **41**:3841–3851 (2025). doi:10.1109/TRO.2025.3577041
23. Levchuk, G., Pattipati, K., Fouse, A., Serfaty, D.: Application of free energy minimization to the design of adaptive multi-agent teams. In: Proceedings SPIE 10206, Disruptive Technologies in Sensors and Sensor Systems, 102060E (2017). doi:10.1117/12.2263542
24. Levchuk, G., Fouse, A., Pattipati, K., Serfaty, D., McCormack, R.: Active learning and structure adaptation in teams of heterogeneous agents. In: Proceedings SPIE 10653, Next-Generation Analyst VI, 1065305 (2018). doi:10.1117/12.2305875
25. Levchuk, G., Pattipati, K., Serfaty, D., Fouse, A., McCormack, R.: Active inference in multiagent systems: Context-driven collaboration and decentralized purpose-driven team adaptation. In: Artificial Intelligence for the Internet of Everything, pp. 67–85. Academic Press, Cambridge, MA (2019). doi:10.1016/B978-0-12-817636-8.00004-1
26. Pezzato, C., Corbato, C. H., Bonhof, S., Wisse, M.: Active inference and behavior trees for reactive action planning and execution in robotics. IEEE Transactions on Robotics **39**(2), 1050–1069 (2023). doi:10.1109/TRO.2022.3226144
27. Baltieri, M., Buckley, C. L.: Nonmodular architectures of cognitive systems based on active inference. In: Proceedings of the 2019 International Joint Conference on Neural Networks, pp. 1–8 (2019). doi:10.1109/IJCNN.2019.8852048

28. Nozari, S., Krayani, A., Marin, P., Marcenaro, L., Gomez, D. M., Regazzoni, C.: Exploring action-oriented models via active inference for autonomous vehicles. EURASIP Journal on Advances in Signal Processing **2024**, 92 (2024). doi:10.1186/s13634-024-01173-9
29. Sawada, H., Ohata, W., Tani, J.: Human-robot kinaesthetic interactions based on the free-energy principle. IEEE Transactions on Systems, Man, and Cybernetics: Systems **55**(1), 366–379 (2025). doi:10.1109/TSMC.2024.3481251
30. Matsumura, T., Esaki, K., Yang, S., Yoshimura, C., Mizuno, H.: Active inference with empathy mechanism for socially behaved artificial agents in diverse situations. Artificial Life **30**(2), 277–297 (2024). doi:10.1162/artl_a_00416
31. Esaki, K., Matsumura, T., Minusa, S., Shao, Y., Yoshimura, C., Mizuno, H., Dynamical perception-action loop formation with developmental embodiment for hierarchical active inference. In: Buckley, C. L., et al. Active Inference. IWAI 2023. Communications in Computer and Information Science 1915, pp. 14–28. Springer, Cham (2023). doi:10.1007/978-3-031-47958-8_2
32. Smets, P., Kennes, R.: The transferable belief model. Artificial Intelligence **66**(2), 191–234 (1994). doi:10.1016/0004-3702(94)90026-4
33. Kullback, S., Leibler, R. A.: On information and sufficiency. The Annals of Mathematical Statistics **22**(1), 79–86 (1951). doi:10.1214/aoms/1177729694